\documentclass[10pt,twocolumn,letterpaper]{article}

\usepackage{iccv}
\usepackage{times}
\usepackage{epsfig}
\usepackage{graphicx}
\usepackage{amsmath}
\usepackage{amssymb}
\usepackage{epstopdf}

 \iccvfinalcopy 

\def\iccvPaperID{1543} 

\ificcvfinal\fi

\begin{document}

\title{Automatic Real-time Background Cut for Portrait Videos}

\author{Xiaoyong Shen \quad Ruixing Wang \quad Hengshuang Zhao \quad Jiaya Jia\\
The Chinese University of Hong Kong\\
{\tt\small \{xyshen, rxwang, hszhao, leojia\}@cse.cuhk.edu.hk}}


\maketitle

\begin{abstract}
We in this paper solve the problem of high-quality automatic real-time background cut for
720p portrait videos. We first handle the background ambiguity issue in semantic
segmentation by proposing a global background attenuation model. A spatial-temporal
refinement network is developed to further refine the segmentation errors in each frame
and ensure temporal coherence in the segmentation map. We form an end-to-end network for
training and testing. Each module is designed considering efficiency and accuracy. We
build a portrait dataset, which includes 8,000 images with high-quality labeled map for
training and testing. To further improve the performance, we build a portrait video
dataset with $50$ sequences to fine-tune video segmentation. Our framework benefits many
video processing applications.
\end{abstract}

\section{Introduction}
Portrait image and video have become conspicuously abundant with the popularity of smart
phones \cite{ShenHJPPSS16}. Portrait segmentation thus plays an important role for
post-processing such as composition, stylization and editing. High performance automatic
portrait video segmentation remains a difficult problem even with recent development on
automatic portrait image segmentation and matting \cite{ShenHJPPSS16,ShenTGZJ16}. We in
this paper tackle this problem starting from following analysis.

\begin{figure}
\centering
\includegraphics[width=0.99\linewidth]{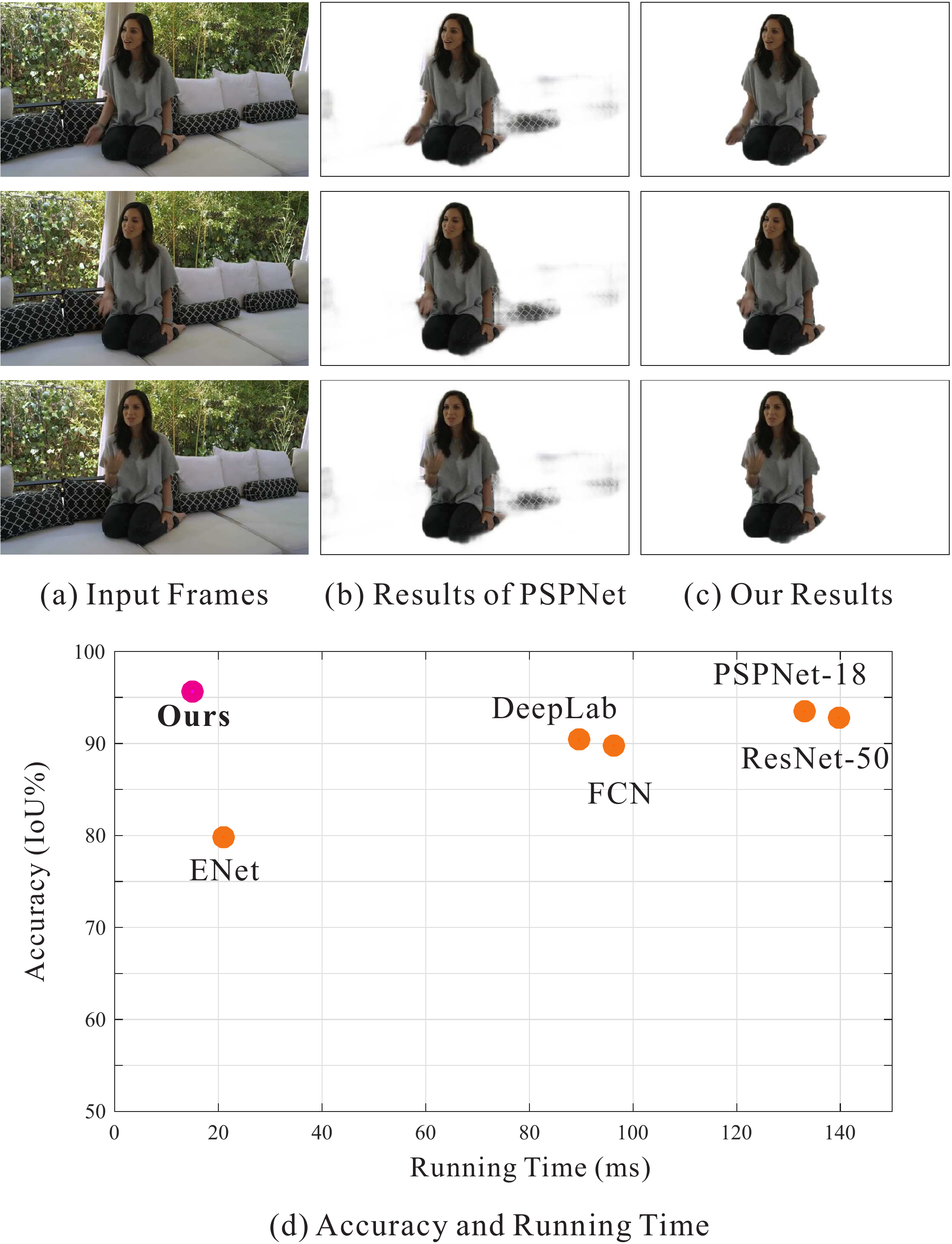}
\caption{Our automatic real-time background cut method.  (a) is input frames in a
portrait video. (b) and (c) show the results of state-of-the-art method \cite{ZhaoSQWJ16}
and ours respectively. (d) shows the accuracy and running time of different segmentation
approaches on our portrait video segmentation dataset.} \label{fig:teaser}
\end{figure}

\vspace{-0.1in}\paragraph{Tedious Interaction Problem} Previous methods
\cite{WangBCAC05,PriceMC09,WangYWZ16,FanZLCC15} need users to specify samples in key
frameworks using strokes and iteratively refine segmentation with more touch-ups, which
are actually labor intensive. We tested the best implementation of Rotobrush in Adobe
After Effect and eventually used one hour to well segment a one-minute video sequence as
shown in Figure \ref{fig:teaser}(a). It would take more time when the video is with more
complicated background or along hair boundaries. Thus, improving segmentation efficiency
is of great practical value for video processing.

\vspace{-0.1in}\paragraph{Time and Accuracy of Automatic Methods} Although many automatic
semantic image segmentation methods \cite{long2015fully,Chen2014_deeplab,ZhaoSQWJ16} were
developed, they are not real-time processing methods even using GPUs for
median-resolution videos, which hinder them from applying to batch and online video
editing. As shown in Figure \ref{fig:teaser}(d), representative methods FCN
\cite{long2015fully}, Deeplab \cite{Chen2014_deeplab}, ResNet \cite{HeZRS16} and PSPNet
\cite{ZhaoSQWJ16} need at least 90ms to process one frame for 720p videos on an Nvidia
Titan X GPU card.  Fast semantic segmentation such as ENet \cite{paszke2016enet}, on the
other hand, can only produce lower-quality results.

The major difficulty of accurate automatic background cut stems from the diverse
complexity of patterns in foreground and background. It is common that background
patterns are brought into foreground estimate even with the powerful PSPNet
\cite{ZhaoSQWJ16} due to the color-texture similarity. One example is shown in Figure
\ref{fig:teaser}(b), where the appearance of the pillow is similar to the foreground
patterns and thus misclassified. Given the very high diversity of background patterns in
indoor and outdoor scenes, it is challenging to reduce this type of misclassification to
a very low level. One focus of this paper is therefore to address this issue.

\vspace{-0.1in}\paragraph{Our Contributions} We propose an automatic real-time portrait
video segmentation method, first addressing foreground and background structure ambiguity
via deep background attenuation, which incorporates extra video background feature
learning to help segmentation. As shown in Figure \ref{fig:teaser}(b) and (c), this
background attenuation scheme can greatly reduce boundary and regional errors.

Second, we design a spatial-temporal video segmentation refinement module to efficiently
improve results by considering video spatial and temporal coherence. Our framework is an
end-to-end trainable convolutional neural network (CNN). We further design Light-ResNet
to achieve the real-time performance, which is over 70 frames/second for testing 720p
videos on an Nvidia Titan X card. State-of-the-art results are achieved in terms of
running time and accuracy as illustrated in Figure \ref{fig:teaser}(d).

To train and evaluate our framework, a portrait segmentation dataset with 8,000 images
whose size are $1200*800$ are built. We also collect a 50 portrait video segmentation
sequences, each with 200 frames. The two datasets are labeled with high-quality
segmentation maps. Our high performance portrait segmentation system can be the
fundamental tool for video processing and benefit all applications requiring high-quality
segmentation maps.

\section{Related Work}

We in this section review the most related image and video segmentation schemes.

\vspace{-0.1in}\paragraph{Graph based Object Segmentation Approaches} Image and video
object segmentation can be a graph-model based problem. For image segmentation, many
methods are based on the graph-cut schemes \cite{Boykov2001interactive}, such as Lazing
Snapping \cite{Li2004_lazy}, Grabcut \cite{Rother2004_grabcut}, and Paint Selection
\cite{Liu2009_paintselection}. These methods need user interaction to specify different
object samples. Besides the graph-cut framework, dense CRF is also applied to object
segmentation as explained in \cite{Krahenbuhl2011_fastcrf}.

Image segmentation can be directly extended to videos by considering the temporal
pixel/object correspondence. Most of the methods pay attention to how to build graph
models \cite{WangTL03,LiuDYO08,YuLZS15}. Approaches of \cite{JangK16,
KunduVK16,GiordanoMPS15,CucchiaraPV03,MaL12,YiP15,Couprie2013,ZhongQP10,ZhangJS13,PerazziWGS15}
introduced different schemes to estimate class object distributions. The geodesic
distance \cite{BaiS09}  was used to model the pixel relation more accurately. To
efficiently solve graph based models, the bilateral space is applied in
\cite{MarkiPWS16}. Energy or feature propagation schemes were also presented in
\cite{Sener2013,ZhuXDYW16}. To reduce user interaction, Nagaraja \etal
\cite{NagarajaSB15} proposed a framework that only needs a few strokes and Lee \etal
\cite{LeeKG11} found key-segments automatically.

Temporal coherence is another important issue in video segmentation. Optical flow
\cite{Tsai0B16,KenderY98}, object/trajectory tracking \cite{FragkiadakiZS12,BrendelT09},
parametric contour \cite{LuBSW16}, long/short term analysis \cite{OchsMB14,JangK16}, \etc
are applied to address the temporal coherence issue. Many previous methods handle bilayer
segmentation \cite{CriminisiCBK06}. Tree-based classifier was presented in
\cite{YinCWE07} and locally competing SVMs were designed in \cite{Gong11} for
better bilayer segmentation. To evaluate video segmentation quality, benchmarks
\cite{PerazziPMGGS16,GalassoNCBS13} were proposed. Compared with these graph based
methods, our method is real-time and without any interaction.

\vspace{-0.1in}\paragraph{Learning based Semantic Segmentation} Previous work focus in
part on learning feature for video segmentation. Price \etal \cite{PriceMC09} learned
multiple cues and integrated them into an interactive segmentation system. Tripathi \etal
proposed learning early- and mid-level features to improve performance. To handle
training data shortage, weakly-supervised and unsupervised learning frameworks were
developed in \cite{WangRLWK16}, \cite{ZhangCLWX15} and \cite{ZhangLL16a} respectively. An
one-shot learning method was proposed in \cite{CaellesMPLCG16} only needing one example
for learning. Drayer \etal proposed a unified framework including object detection,
tracking and motion segmentation for object-level segmentation. To reduce errors during
propagation, Wang \etal \cite{WangYWZ16} developed segmentation rectification via
structured learning.

In recent years, CNNs have achieved great success in semantic image segmentation.
Representative work exploited CNNs in two ways. The first is to learn important features
and then apply classification to infer pixel labels
\cite{Arbelaez2012,Mostajabi2014,Farabet2013}. The second way is to directly learn the
model from images. Long \etal \cite{long2015fully} introduced fully convolutional networks.
Following it, DeepLab \cite{Chen2014_deeplab} and CRFasRNN \cite{Zheng2015_crfrnn} were
developed using CRF for label map refinement. Recent PSPNet \cite{ZhaoSQWJ16} is based on
ResNet \cite{HeZRS16}, which performs decently.

These frameworks can be directly applied to videos in a frame-by-frame fashion. To
additionally deal with temporal coherence, spatial-temporal FCN \cite{FayyazSSFK16} and
recurrent FCN \cite{ValipourSJR16,SiamVJR16,NilssonS16} were proposed. Shelhamer \etal
\cite{ShelhamerRHD16} proposed Clockwork Convnets driven by fixed or adaptive clock
signals that schedule processing of different layers. To use the temporal information,
Khoreva \etal \cite{KhorevaPBSS16} predicted per-frame segmentation guided by the output
of previous frameworks. These approaches aim at general object segmentation. They have
difficulty to achieve real-time performance for good quality portrait video segmentation.

\vspace{-0.1in}\paragraph{Video Matting Schemes} Similar to image matting, video matting
computes the alpha matte in each frame. A survey of matting techniques can be found in
\cite{WangC07} and an evaluation benchmark is explained in \cite{ErofeevGVFW15}. Most
video matting methods extend the image one by adding temporal consistency. Representative
schemes are those of
\cite{ZouCCW15,ShahrianPCR14,LiCT13,ChoiLT12,BaiS09,ApostoloffF04,ChuangACSS02}. Since
the matting approaches need user specified trimaps, methods of \cite{JuWLWD13,GongQC15}
applied segmentation to improve trimap quality. Our method automatically achieves
portrait segmentation and generates trimaps for further video matting.

\section{Our Framework}

\begin{figure}
\centering
\includegraphics[width=0.99\linewidth]{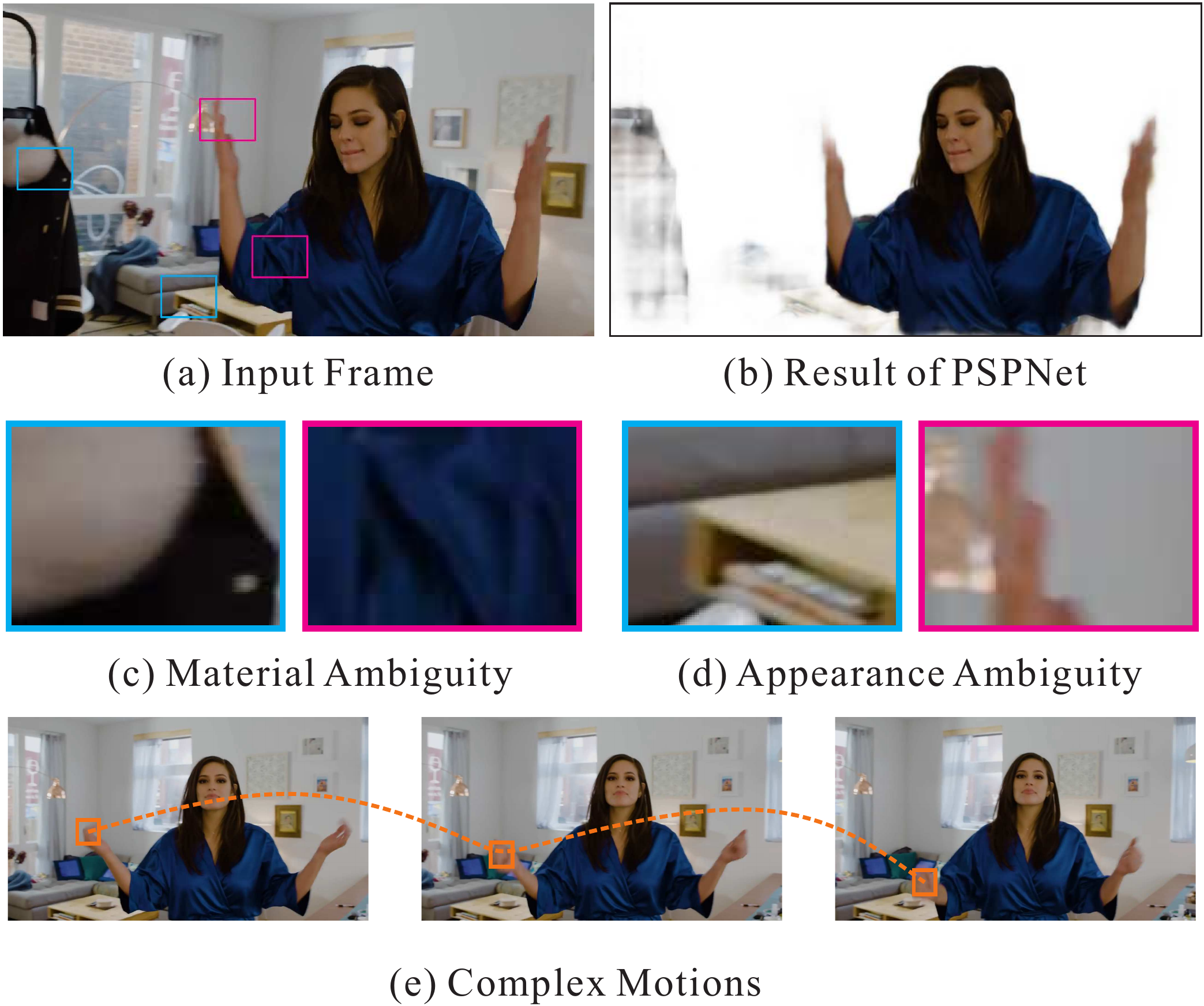}
\caption{Difficulty of portrait video segmentation. (a) is an input frame and (b) shows
the segmentation result by state-of-the-art method \cite{ZhaoSQWJ16}. (c) and (d) show
the ambiguities stemming from material and appearance similarity. (e) shows the complex
motion in portrait videos. } \label{fig:diffs}
\end{figure}

\begin{figure*}
\centering
\includegraphics[width=0.99\linewidth]{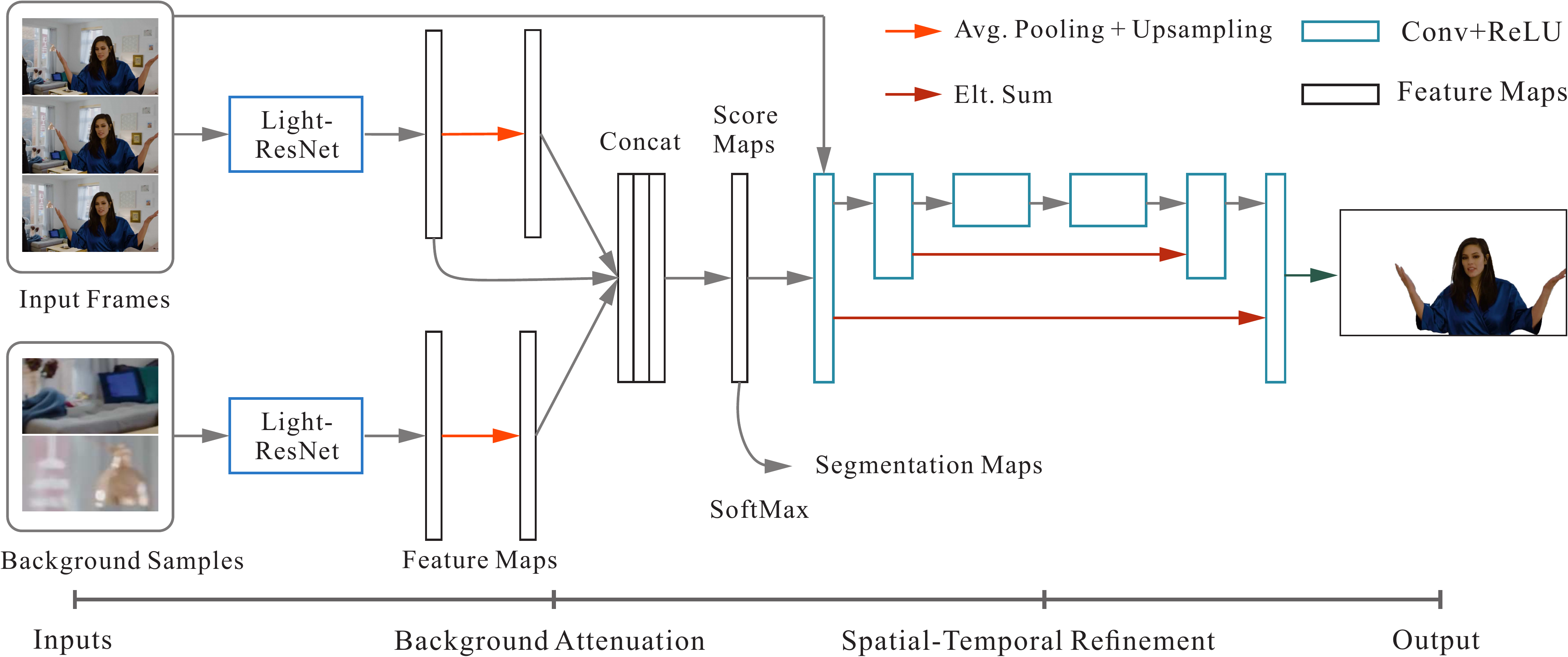}
\caption{Our automatic high-quality real-time portrait video segmentation framework.} \label{fig:framework}
\end{figure*}

Our end-to-end trainable framework is illustrated in Figure \ref{fig:framework}. It
addresses two main challenges. The first is the ambiguity between foreground and
background patterns. As shown in Figure \ref{fig:diffs}, material and appearance in (c)
and (d) are very similar, making even state-of-the-art semantic segmentation method
\cite{ZhaoSQWJ16} fail to cut out foreground accurately. We design a deep background
attenuation model to address this challenge.

The second challenge is on complex motion as shown in Figure \ref{fig:diffs}(e) that may
cause correspondence estimation to fail. Also, fast motion could  blur the content. We
address this challenge with a spatial-temporal refinement module. These modules are
implemented with the in-depth consideration of short running time and high quality.

As shown in Figure \ref{fig:framework}, our framework takes successive $2n+1$ frames
$\{I^{t-n}...I^{t+n}\}$ as input and outputs the segmentation map of $I^t$ ,where $I^t$
denotes the $t$th frame of the video.

\subsection{Global Background Attenuation}

Our global background attenuation is to collect a few background samples around the same
scene without the requirement of alignment or fixing cameras, and use them to globally
attenuate background pixels. It is a rather easy setting taking a few seconds prior scene
capture for following live video processing. The background samples can be also got by 
manually cropping out the video background.  It is different from the method of
\cite{SunZTS06} where the latter requires stationary background.

\vspace{-0.1in}\paragraph{Segmentation with Global Attenuation} After collecting a few
frames shooting the background from arbitrary locations and angles, we process them
through a network. As shown in Figure \ref{fig:framework}, our overall network includes
two paths. The first extracts feature maps of the input frames $\{I^{t-n}...I^{t+n}\}$
and the second path computes features of the background samples. Both of branches are
specially designed light-weight ResNet, which we will detail later. Similar to previous
segmentation frameworks \cite{Chen2014_deeplab,ZhaoSQWJ16}, the extracted feature maps in
our first path are further processed by one convolution layer to obtain the score maps,
corresponding to the segmentation result after Softmax operation as illustrated in Figure
\ref{fig:framework}.

Our second path plays the role of background attenuation. Since these background samples
are not aligned with the input frames, directly apply extracted background feature to
attenuate the segmentation feature maps is vulnerable to errors caused by object/region
discrepancy. We address this issue by estimating the global background features. It
starts from the extracted background feature by adding global average pooling and
upsampling. The upsampling step makes global feature map keep the same size as the
inputs. The final background global features are concatenated with the segmentation
feature maps from the first path as illustrated in Figure \ref{fig:framework}.

The features adopted in previous segmentation frameworks \cite{LiuRB15,ZhaoSQWJ16} are
used to enlarge the receptive fields. They are similar to our first path. In contrast,
the way to extract global features for background samples for our special task of
attenuation is new and empirically effective.

\begin{figure}
\centering
\includegraphics[width=0.99\linewidth]{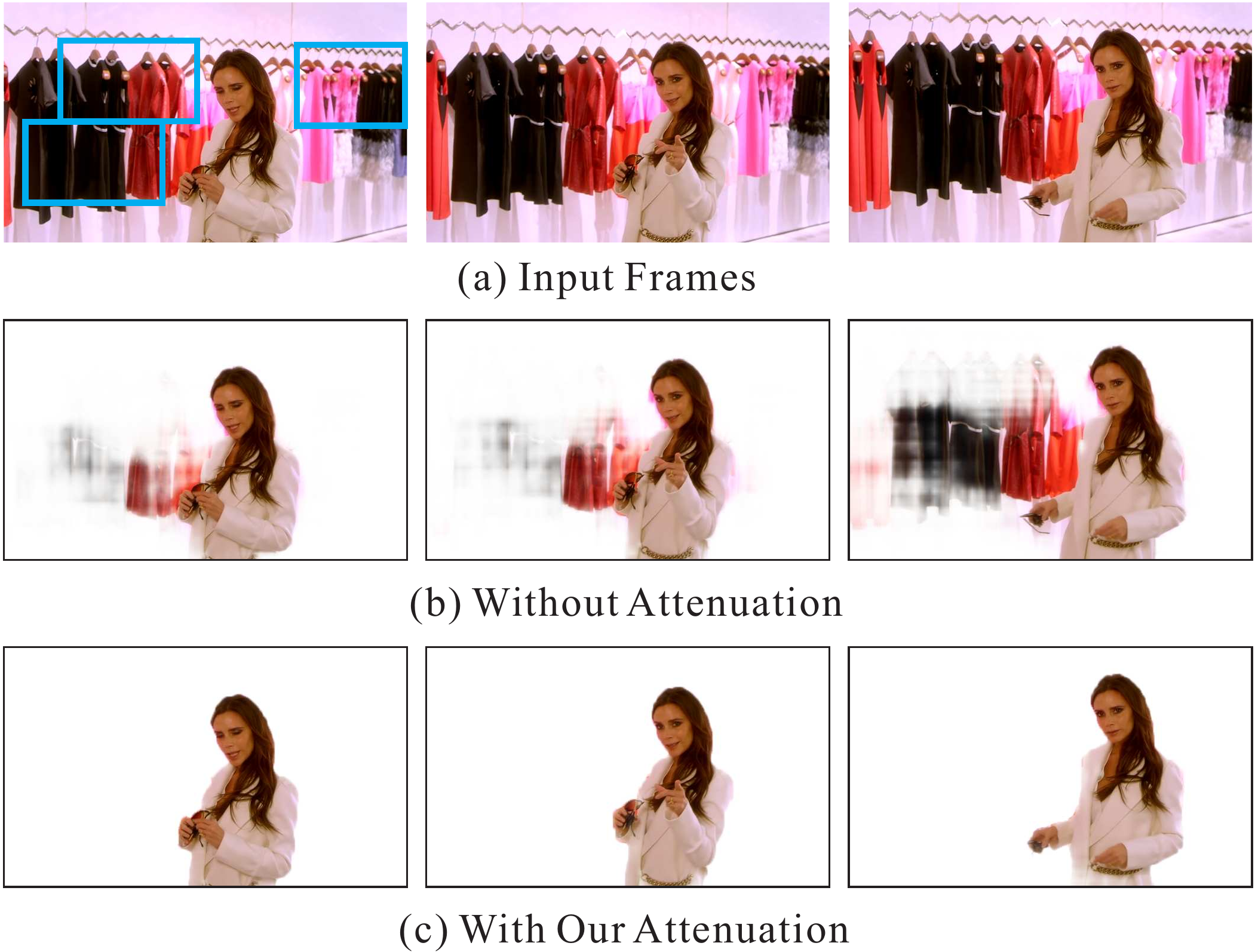}\\
\caption{Effectiveness of our background attenuation. (a) Input frames. (b)-(c) Results
without and with background attenuation respectively. Background samples are inside
rectangles in (a).} \label{fig:attenuation}
\end{figure}

\vspace{-0.1in}\paragraph{Effectiveness of Our Attenuation} Our background attenuation
can quickly reduce segmentation errors. To demonstrate it, we shown an example in Figure
\ref{fig:attenuation} where foreground and background both have clothes for segmentation.
Directly applying the segmentation CNNs in our upper branch in Figure \ref{fig:framework}
cannot estimate correct foreground -- results are shown in (b). With our background
attenuation, the network outputs much higher quality results as shown in (c) and the
background samples are shown in (a) in the highlighted rectangles. Note that in this
example the video background is not stationary and not aligned with input images. Yet its
usefulness in our framework is clearly exhibited.

\vspace{-0.1in}\paragraph{Light-ResNet for Real-time Performance} To achieve real-time
performance, we designed a Light-ResNet to balance accuracy and efficiency as illustrated
in Figure \ref{fig:framework}. Our network is based on ResNet-18 \cite{HeZRS16} and
goes through model compression to accelerate the network. Since the running time
bottleneck is mostly related to the large number of output channels of convolution
layers, we prune convolution filters in each layer \cite{LiKDSG16}. We gradually reduce
the low-response filters and fine-turn the compressed model. For each step, we keep 90\%
of filters based on previous fine-turned model and finally retain only 20\% of filters
compared to the original ResNet-18 model. Our system only needs 14.99ms to test one 720p
color image, while the original structure takes 110.91ms. The accuracy is not much
sacrificed on our testing dataset, discussed later in experiments. We note that gradually
pruning filters is essential for balancing performance and speed.

\subsection{Spatial-temporal Refinement}
To enforce temporal coherence and deal with remaining spatial segmentation errors, we
propose the spatial-temporal refinement network as shown in Figure \ref{fig:framework} to
further improve result quality.

\vspace{-0.1in}\paragraph{Network Design} In our refinement network, the input is the
segmentation score maps of the input frames $\{I^{t-n}...I^{t+n}\}$. The original images
are taken as guidance for refinement. The output is updated segmentation of frame $I^t$
as shown in Figure \ref{fig:framework}. We first utilize three convolution layers to
extract spatial-temporal features \cite{long2015fully,RonnebergerFB15}. Considering
inevitable displacement, we use stride 2 for downsampling to reduce this effect. Then
three corresponding deconvolution layers are employed to upsample previously downsampled
feature maps. We further fuse the same-size feature maps between feature extraction and
deconvolution layers by summation to improve results (Figure \ref{fig:framework}).

Compared with previous refinement network \cite{LiaoTLMJ15}, which did not apply
downsampling or pooling operations to preserve image details, ours applies sampling to
address the displacement problem and accelerate computation. Moreover, our network does
not need to regress image details.

\begin{figure}
\centering
\includegraphics[width=0.99\linewidth]{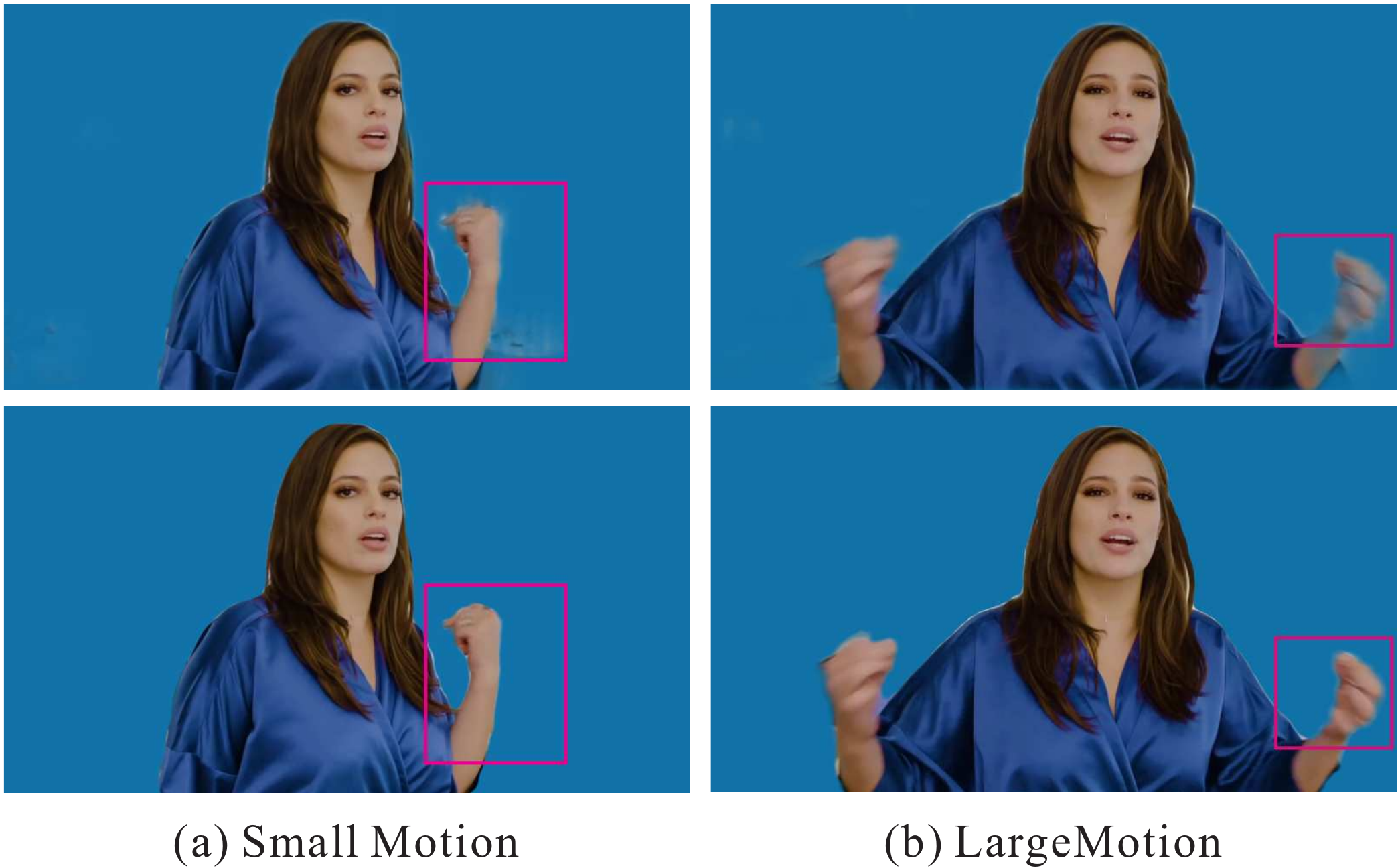}\\
\caption{Effectiveness of refinement using our spatial-temporal networks. (a) shows the
refinement on a small motion case and (b) is with large motion. The top row shows results
without refinement and the bottom includes our final results. } \label{fig:refinement}
\end{figure}

\vspace{-0.1in}\paragraph{Analysis of the Refinement} Our spatial-temporal refinement can
further uplift edge accuracy and reduce small-segment errors as shown in Figure
\ref{fig:refinement}. For small-motion cases shown in (a), our refinement sharps edges
for better quality. Our network has the reasonable ability to handle large motion as
shown in Figure \ref{fig:refinement}(b). The reason is that the input color images are
applied as guidance in our network for it to learn the way to combine temporal and
spatial information.

\section{Model Implementation}

Our network can be trained via an end-to-end scheme. In addition to the details below,
more are included in our supplementary file.

\subsection{Data Preparation}
We prepare two datasets to train our model. The first is the portrait image segmentation
dataset, which includes 8,000 portrait images, each with labeled masks. The second
dataset is with 50 portrait sequences, each with 200 frames and labels. For the first
dataset, we split the data into training and testing sets with 7,670 and 330 images
respectively. We also use 45 portrait videos in training and 5 in testing.

\begin{figure}[t]
\centering
\includegraphics[width=0.99\linewidth]{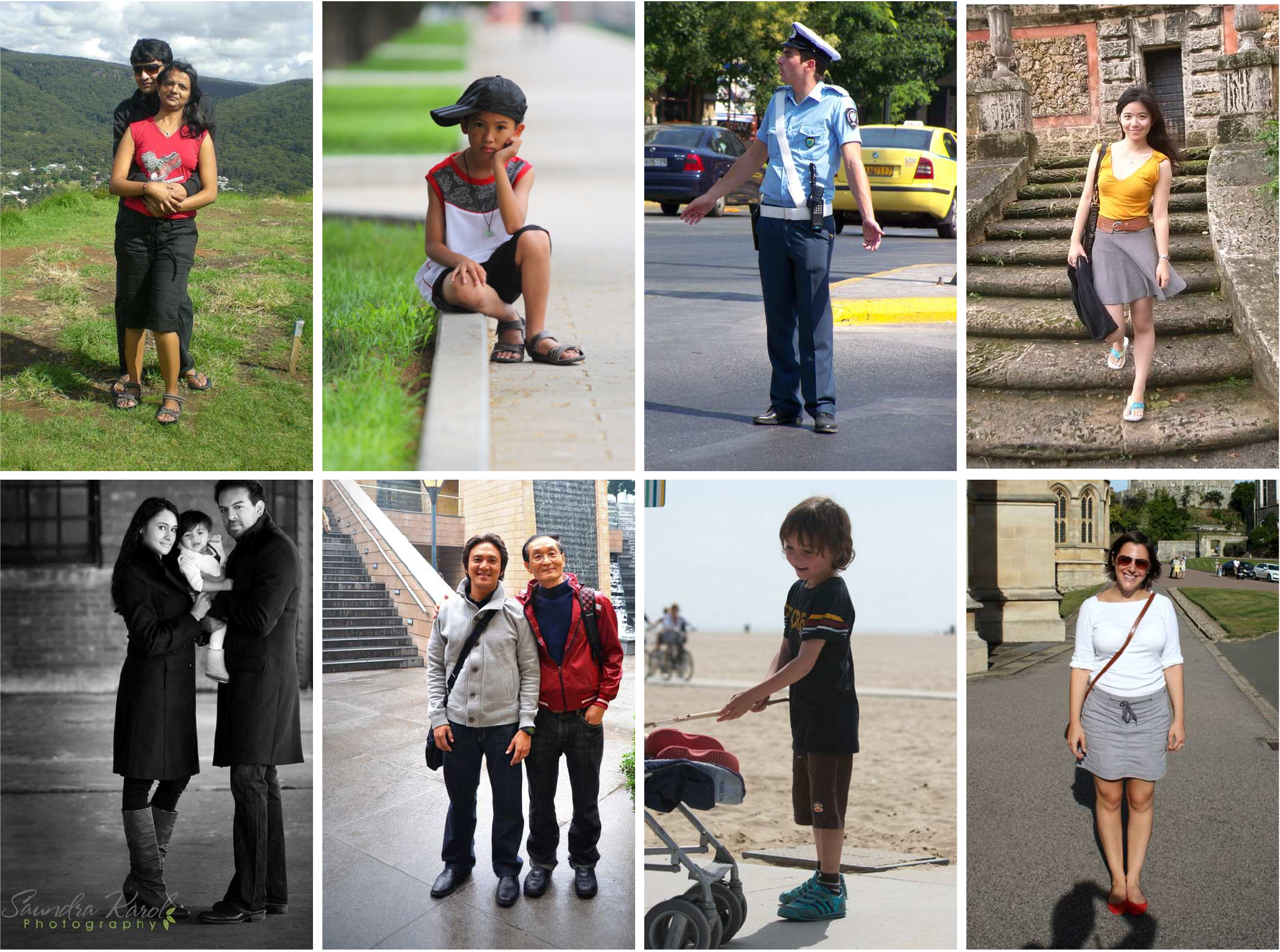}
\caption{Data examples in our portrait dataset.} \label{fig:dataexample}
\end{figure}

Examples of our portrait images are shown in Figure \ref{fig:dataexample}. We first
collect a large number of images from Flickr by searching keywords
``portrait'',``human'', ``person'', \etc Then we process each downloaded image using the
person detector \cite{RenHGS15} to crop out persons and adjust each image to resolution
$1200\times800$ as shown in Figure \ref{fig:dataexample}. Finally, we consider the
variety in person age, appearance, pose, accessories \etc and keep the most diverse 8,000
portraits in our final dataset. To get the segmentation ground truth, we label each
portrait by the Adobe Photoshop quick selection tool.

The portrait videos are from 5 different places with 10 people inside. Complex poses are
required for each person. We also captured the background samples for our model training
-- each video is only with 20 frames of background {\it not} aligned with any following
person-involved frames. We labeled each video using Adobe After Effect's Rotobrush by
iteratively refining each frame. Examples are included in our supplementary file.

The reason that collect two datasets is to cover diverse persons and have data to learn
temporal coherence. We note it is still very difficult to collect and label a large
portrait video dataset covering a greater quantity of persons.

\subsection{Training and Testing Details}
We train our model using the two datasets and implement our model using Caffe
\cite{jia2014caffe} with one Nvidia Titan X GPU card. We train our Light-ResNet using the
7,670 portraits. We only apply the Softmax loss similar to semantic segmentation
\cite{ZhaoSQWJ16}. During training, the batch size is set to 16 and each sample is
randomly cropped to $569\times569$. The initial learning rate is set to $1e-3$. We change
it using the ``poly'' policy with gamma 0.9. 40 epochs are conducted for our model
training.

Then, we train the whole network by the portrait video data -- the 45 sequences include
9,000 frames and also corresponding separate unaligned background samples. The two
Light-ResNet modules are initialized by our previously trained Light-ResNet model. The
Softmax loss is added similar to the first step and the $L_2$ loss is added for the
refinement network. For each input frame, we randomly select one background sample and we
set the batch size to 16 and $n$ to 2 -- it means we use five frames for each-pass
spatial-temporal refinement during training. The learning rate in the refinement network
is set to 10 times of the Light-ResNet because no pre-trained weights are provided. The
base learning rate and the changing policy remain the same.

To test our model, we first pre-compute the background global features using the provided
background samples. Then we test the neighboring five frames as a batch using the
background attenuation model and finally we apply the spatial-temporal refinement via
sliding windows. This scheme makes each frame only needs to be computed once. For the
720p video, our whole network only takes 17.57ms where 14.99ms is for the background
attenuation segmentation model and 2.58ms is for spatial-temporal refinement. All our
experiments are conducted with one graphics card as explained above.

\section{Evaluation and Applications}

We in this section evaluate our methods and also show the applications.

\subsection{Evaluations and Comparisons}

\begin{table}[t]
\centering
\small
\begin{tabular}{l|c}
  \hline
  Methods & Accuracy (Mean IoU\%)\\
  \hline
  \hline
  FCN \cite{long2015fully}          &91.62 \\
  DeepLab \cite{Chen2014_deeplab}   &91.59 \\
  PSPNet-50 \cite{ZhaoSQWJ16}       &93.51 \\
  PSPNet-18 \cite{ZhaoSQWJ16}       &93.33 \\
  ENet \cite{paszke2016enet}        &82.58 \\
  Ours w/o Atten. w/o Refin.              &93.04 \\
  Our with background training &     94.13        \\
  Ours with Atten. w/o Refin.              &96.49 \\
  \hline
  \textbf{Ours}                  &\textbf{96.74} \\
  \hline
\end{tabular}\vspace{0.1in}
\caption{Comparisons of different video segmentation methods on our portrait video dataset.}\label{tab:segonvideo}
\end{table}

\begin{figure}[t]
\centering
\includegraphics[width=0.99\linewidth]{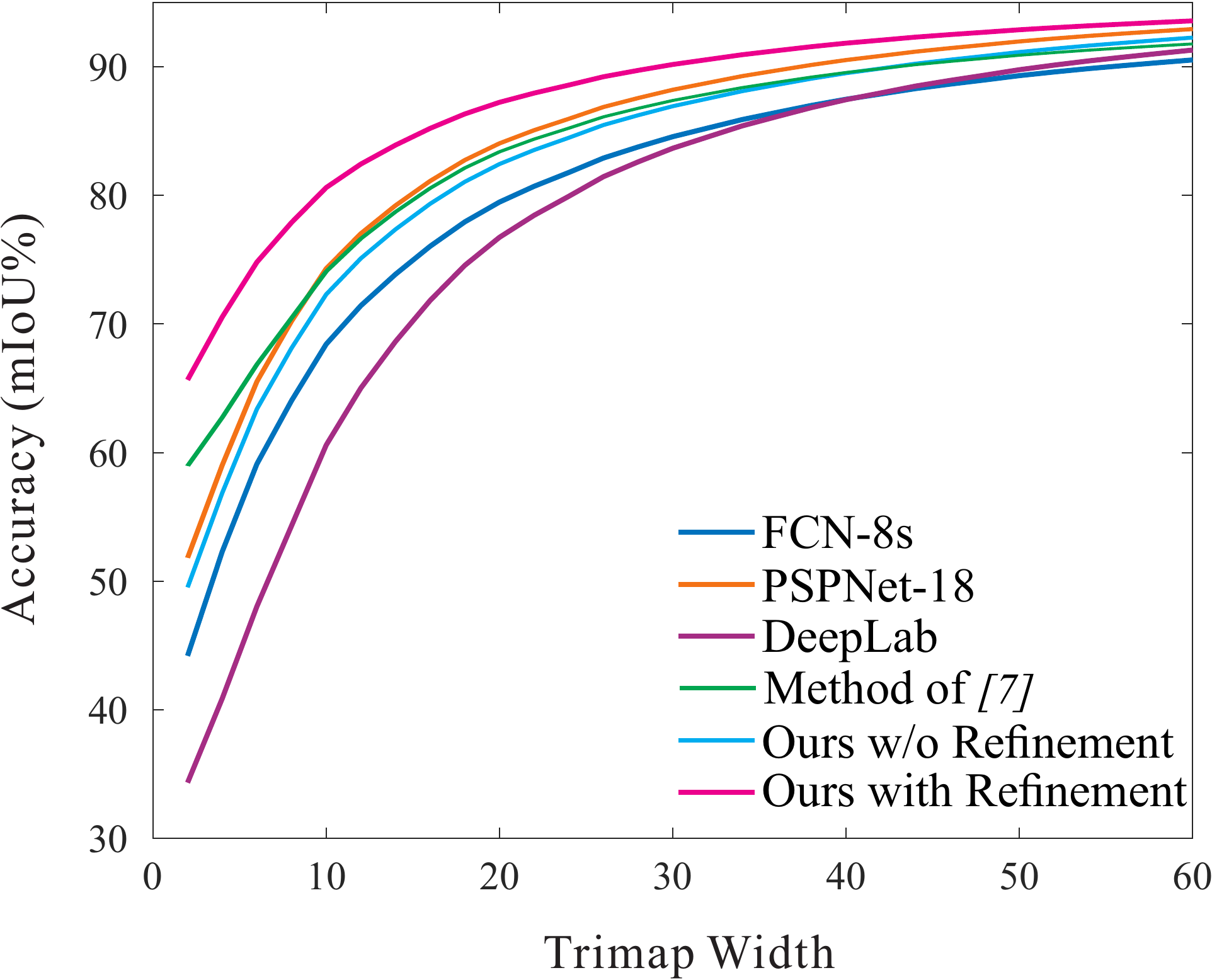}
\caption{Comparisons of boundary accuracy of different methods. } \label{fig:boundary}
\end{figure}

\vspace{0.1in}\noindent\textbf{Portrait Video Segmentation Evaluation~~} We first compare
different methods on our portrait video segmentation dataset. The mean
Interaction-over-Union (IoU) metric is applied to measure accuracy on our 1,000 video
frames on our testing data. We compare our method with the representative semantic
segmentation methods of FCN \cite{long2015fully}, DeepLab \cite{Chen2014_deeplab}, PSPNet
\cite{ZhaoSQWJ16} and ENet \cite{paszke2016enet}. For each method, we re-train the
portrait segmentation model and fine turn it from corresponding public models.

The number of output channels is set to 2 for the two-class segmentation problem. All
models are trained with our portrait image segmentation and portrait video segmentation
datasets that contain 7,670 portrait images and 9,000 video frames respectively. The
training policy follows the original paper. We test each method on our video testing
sequences frame-by-frame and the results are reported in Table \ref{tab:segonvideo}.

Among all these methods, state-of-the-art semantic segmentation PSPNet \cite{ZhaoSQWJ16}
achieves the best performance on our portrait video segmentation dataset, complying with
results on general object segmentation on datasets of ImageNet and Cityscapes. Because of
simplicity of the model, ENet \cite{paszke2016enet} does not perform similarly well.

\begin{table}[t]
\centering \small
\begin{tabular}{l|c|c}
  \hline
  Methods & Accuracy (Mean IoU\%) & Time (ms)\\
  \hline
  \hline
  FCN \cite{long2015fully}          &89.68      &94.25\\
  DeepLab \cite{Chen2014_deeplab}   &89.72      &90.61\\
  PSPNet-50 \cite{ZhaoSQWJ16}       &93.56      &139.7\\
  PSPNet-18 \cite{ZhaoSQWJ16}       &92.81      &110.9\\
  ENet \cite{paszke2016enet}        &79.86      &21.00\\
  \hline
  Our Light-ResNet                  &91.34      &14.99\\
  \hline
\end{tabular}\vspace{0.1in}
\caption{Evaluations of the Light-ResNet on our portrait segmentation
dataset.}\label{tab:segonportrait}
\end{table}

We then evaluate our method based on results reported in Table \ref{tab:segonvideo}.
Our system achieved the best performance. High importance of background attenuation and
spatial-temporal refinement is also revealed from Table \ref{tab:segonvideo}. The
background attenuation can greatly improve the quality because it can reduce errors
caused by the ambiguity between foreground and background.

We note that directly adding the background images for model training can only yield
improvement of about 1\% IoU. Compared with our attenuation that improves 3.7\% IoU, the
straight-forward background sample training is obviously not optimal. Our
spatial-temporal refinement further improve the accuracy. Since the improvement is mainly
on edges, it cannot be accurately measured by IoU and we analyze it more below.

\begin{figure*}[t]
\centering
\includegraphics[width=0.95\linewidth]{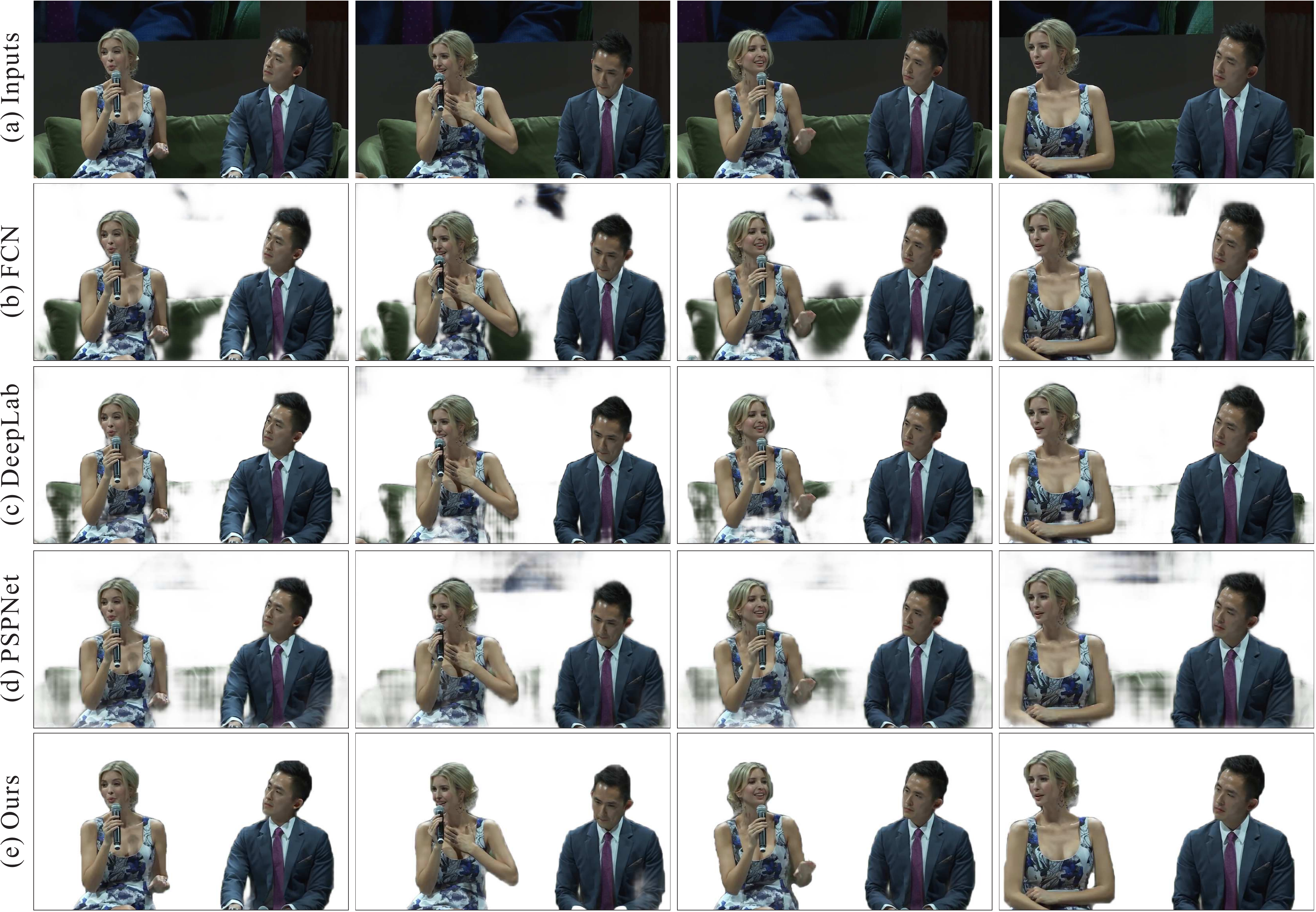}
\caption{Comparison of different methods. (a) is the input and (b) is the result of FCN
\cite{long2015fully}. (c) is the result from DeepLab \cite{Chen2014_deeplab} and (d) is
computed by PSPNet \cite{ZhaoSQWJ16}. (e) is ours. } \label{fig:comparison1}
\end{figure*}

\begin{figure*}[t]
\centering
\includegraphics[width=0.96\linewidth]{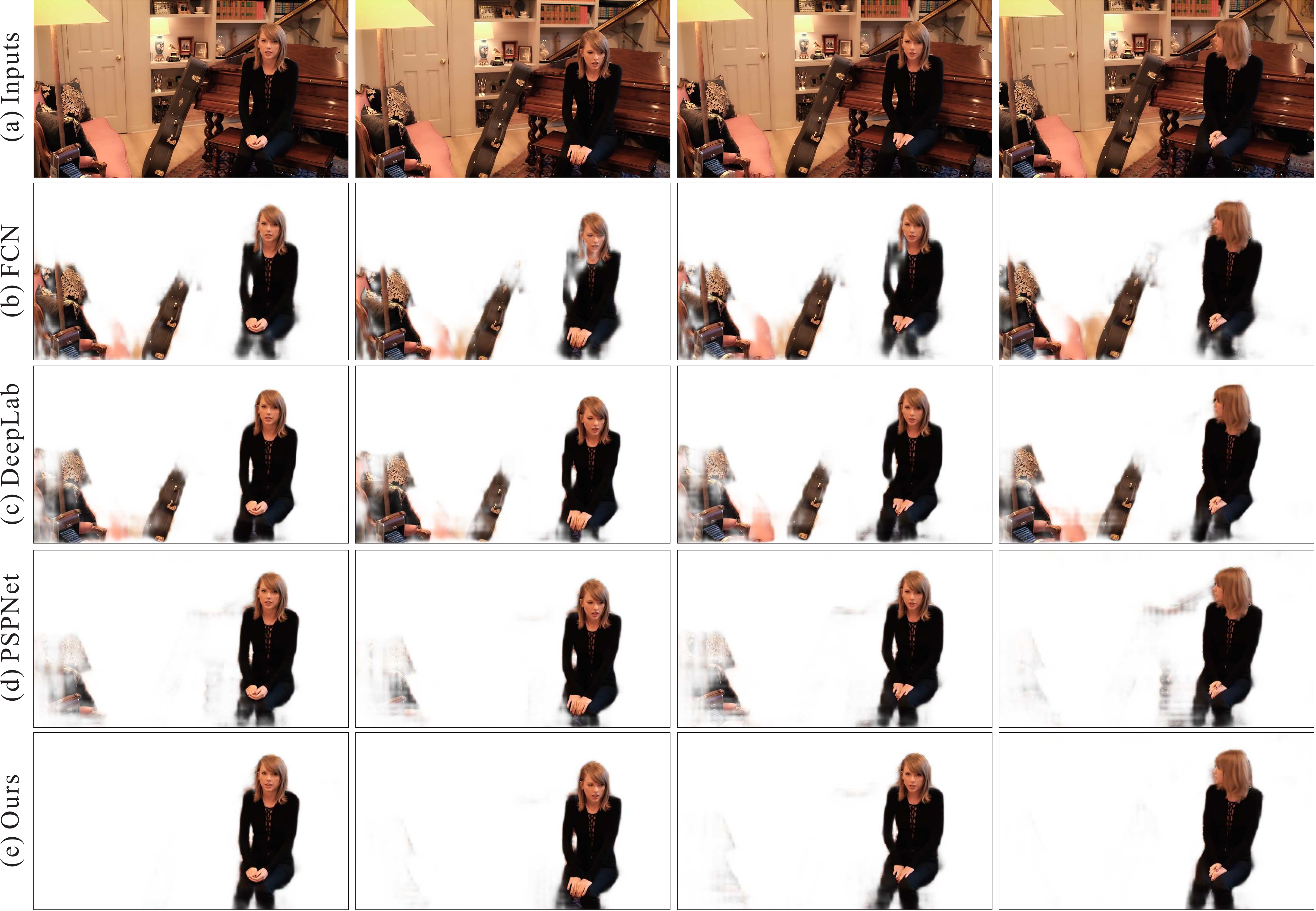}
\caption{More comparisons. (a) is the input and (b) is the result of FCN
\cite{long2015fully}. (c) is the result from DeepLab \cite{Chen2014_deeplab} and (d) is
computed by PSPNet \cite{ZhaoSQWJ16}. (e) is ours.} \label{fig:comparison2}
\end{figure*}

\vspace{0.1in}\noindent\textbf{Effectiveness of the Spatial-temporal Refinement~~} Our
spatial-temporal refinement plays an important role to improve boundary accuracy for
video segmentation. To evaluate it, we compute the IoU only near the ground truth
boundary. Similar to \cite{ChenBP0Y16}, we generate different width trimaps centered at
the ground truth boundary and then compute the mean IoU only on the trimap in our
portrait video segmentation testing datasets. The changes between the mean IoU and the
trimap width of different methods are shown in Figure \ref{fig:boundary}.

Since the methods FCN \cite{long2015fully}, DeepLab \cite{Chen2014_deeplab}, PSPNet
\cite{ZhaoSQWJ16} and ours without refinement do not have special boundary post-processing, the boundary
accuracy is not that satisfying. Method of \cite{ChenBP0Y16} learns domain transform
filter to refine the score map in the network and thus improves boundary accuracy. As
shown in Figure \ref{fig:boundary}, our method with spatial-temporal refinement can
effectively improve accuracy near boundary.

\vspace{0.1in}\noindent\textbf{Evaluation of the Light-ResNet for Portraits~~} We compare
our Light-Resnet for segmentation with representative segmentation schemes of FCN
\cite{long2015fully}, DeepLab \cite{Chen2014_deeplab}, PSPNet \cite{ZhaoSQWJ16} and ENet
\cite{paszke2016enet}. Similar to the evaluation in portrait video segmentation dataset,
we first update the network to binary-label output and use our portrait image dataset to
train it. Each method is trained using the originally released code following authors
instruction.

The accuracy and running time of each method are reported in Table
\ref{tab:segonportrait}. Methods of FCN and DeepLab take over 90ms to test an image and
the mean interaction-over-union (IoU) is near 90\%. Although the PSPNet achieved the best
performance, its running time is over 110ms. The ENet is very efficient; but the
segmentation accuracy is an issue. As shown in Table \ref{tab:segonportrait}, our method
only takes 14.99ms to test a color image with size $1200\times800$ and the accuracy is
comparable to state-of-the-art PSPNet on our portrait image testing dataset.

\begin{figure*}
\centering
\begin{tabular}{@{\hspace{0.0mm}}c@{\hspace{1.0mm}}c@{\hspace{1.0mm}}c@{\hspace{1.0mm}}c@{\hspace{1.0mm}}c@{\hspace{0mm}}}
\includegraphics[width=0.19\linewidth]{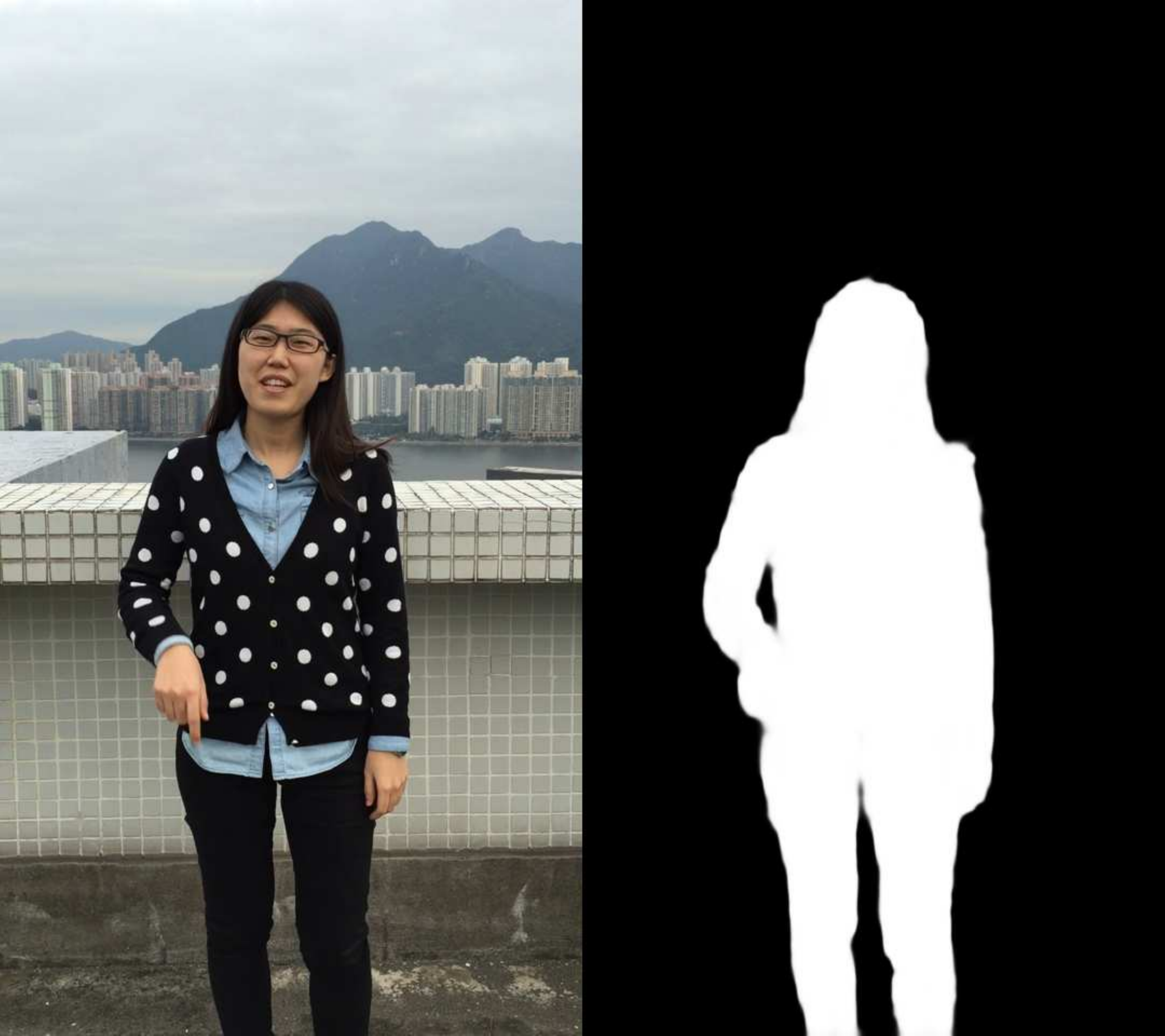}&
\includegraphics[width=0.19\linewidth]{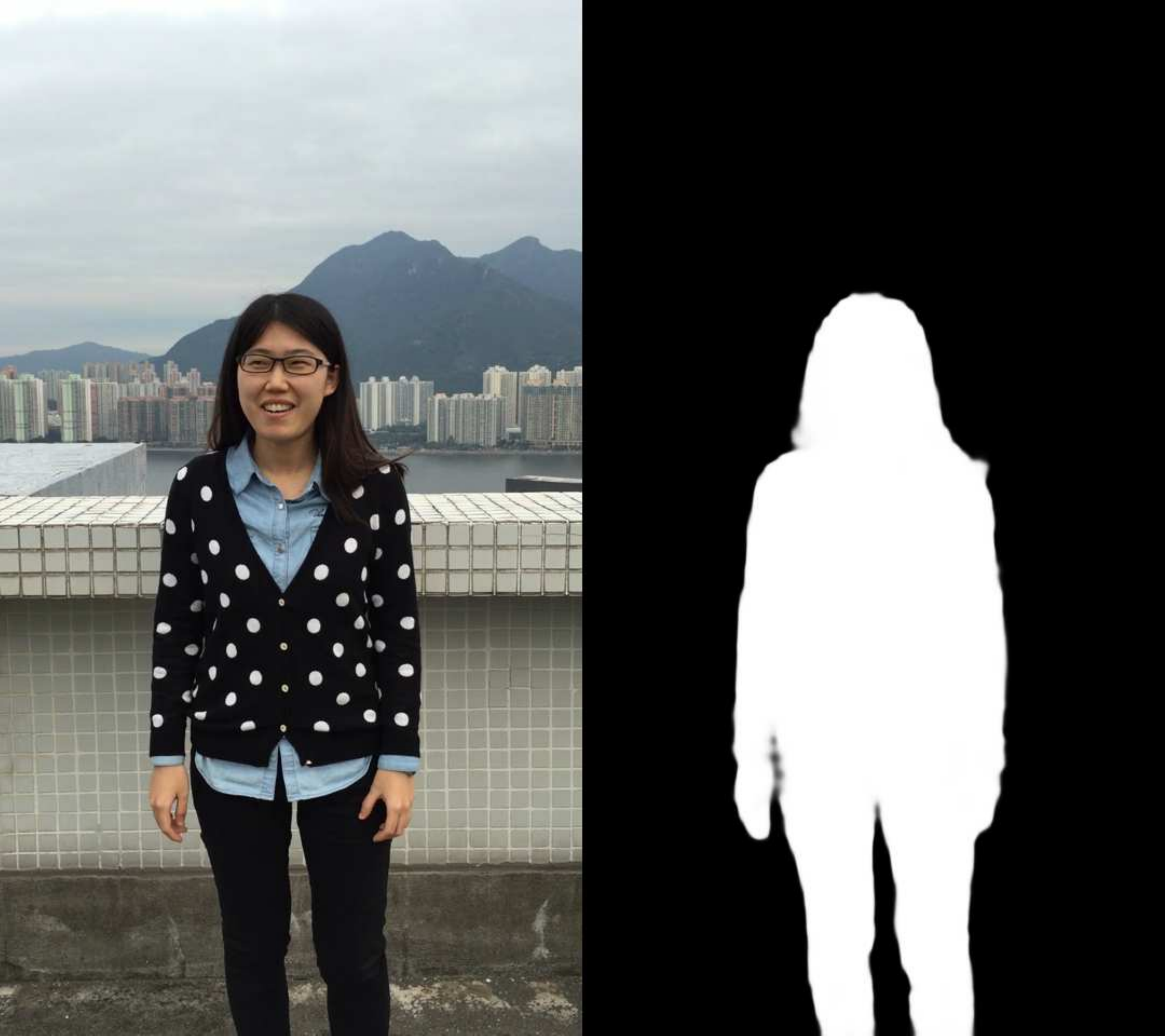}&
\includegraphics[width=0.19\linewidth]{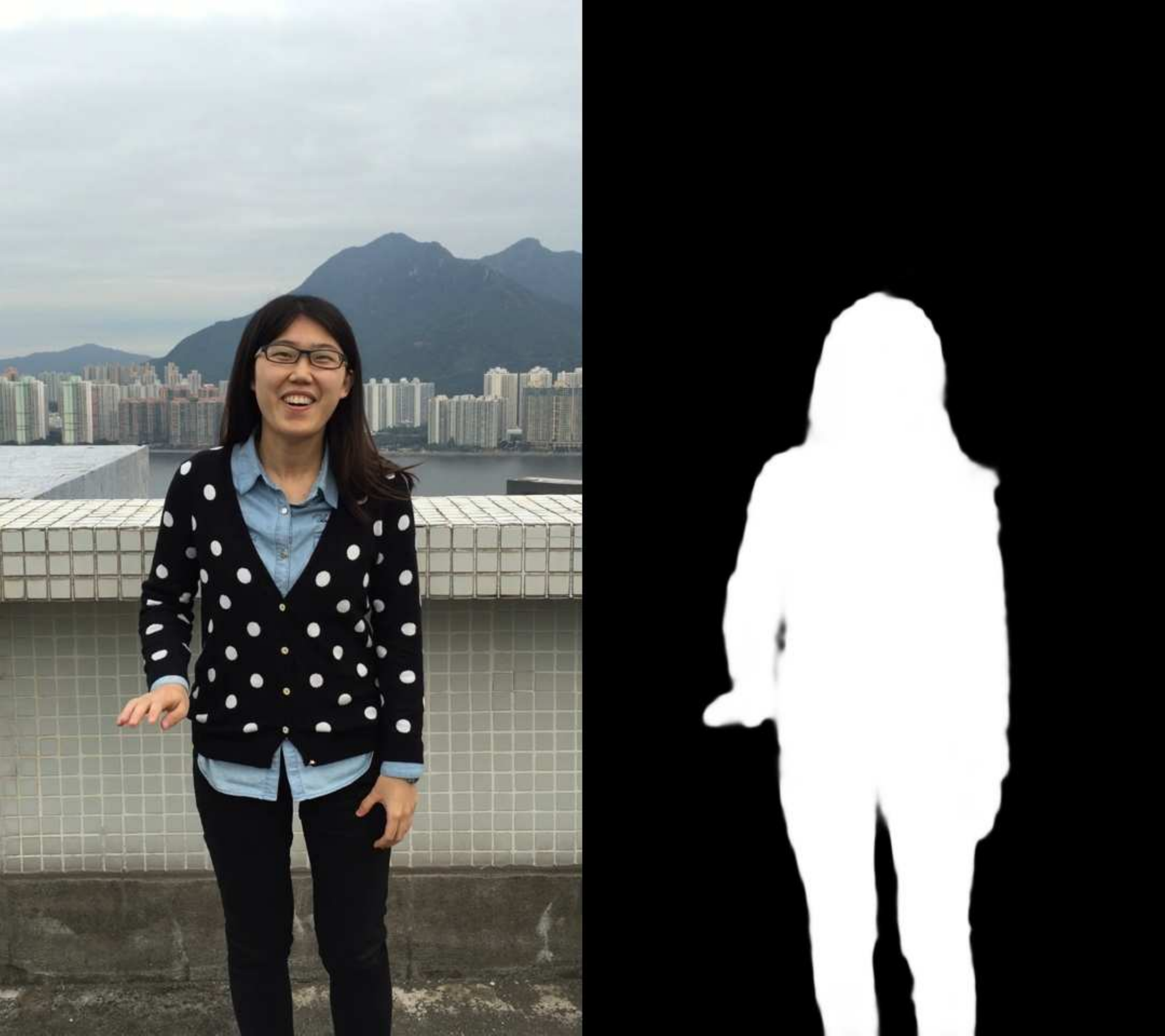}&
\includegraphics[width=0.19\linewidth]{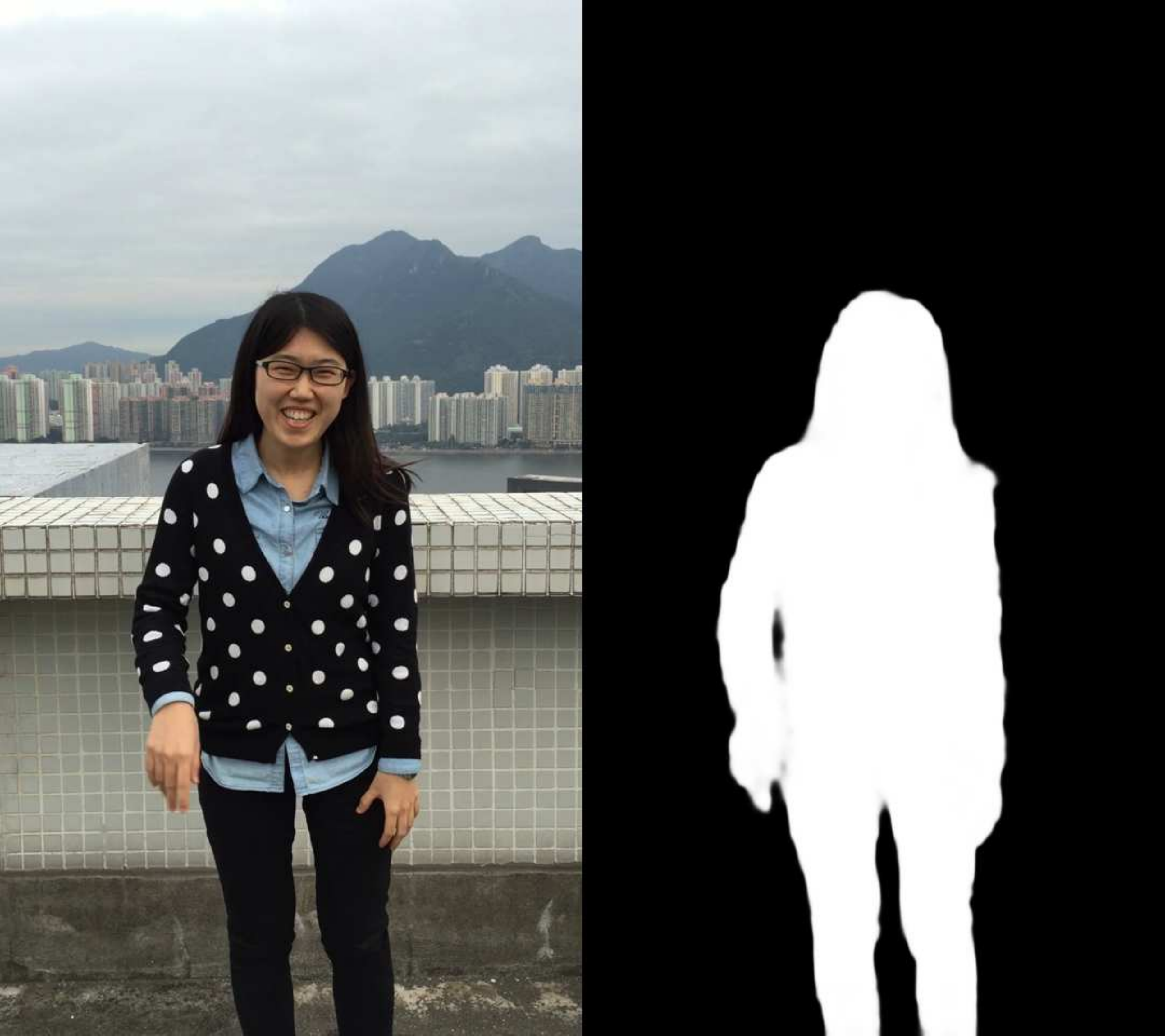}&
\includegraphics[width=0.19\linewidth]{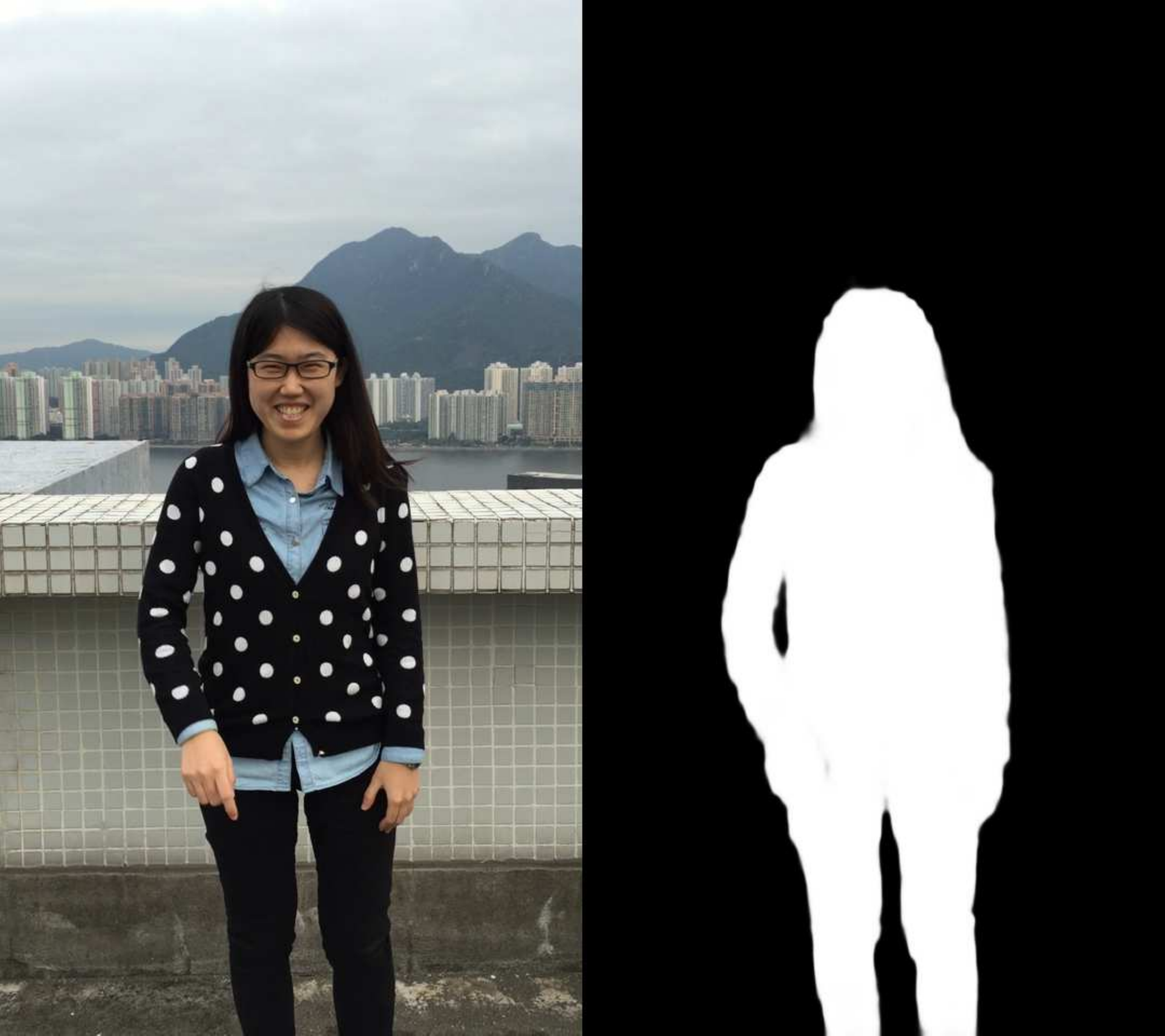}\\
\includegraphics[width=0.19\linewidth]{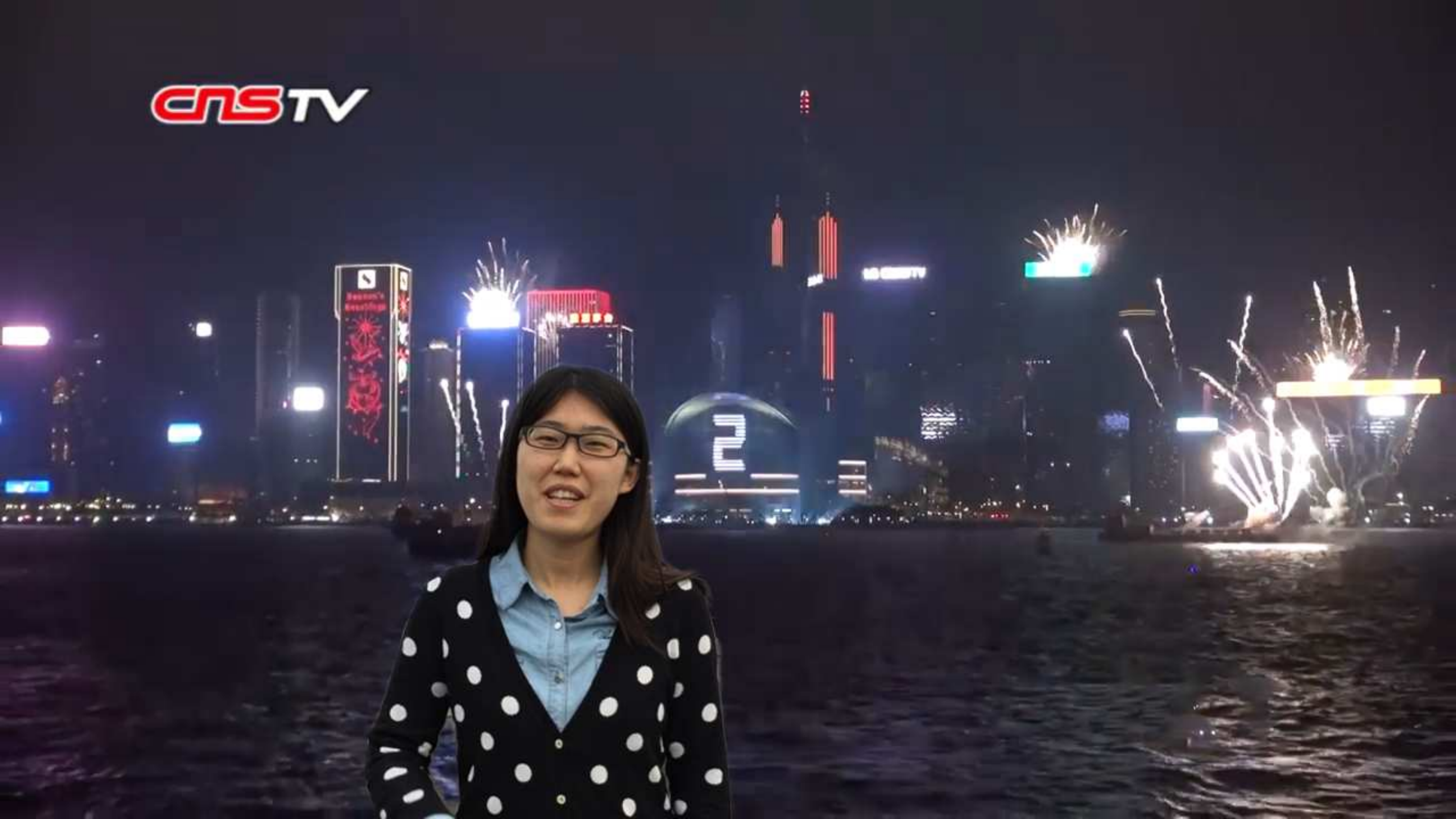}&
\includegraphics[width=0.19\linewidth]{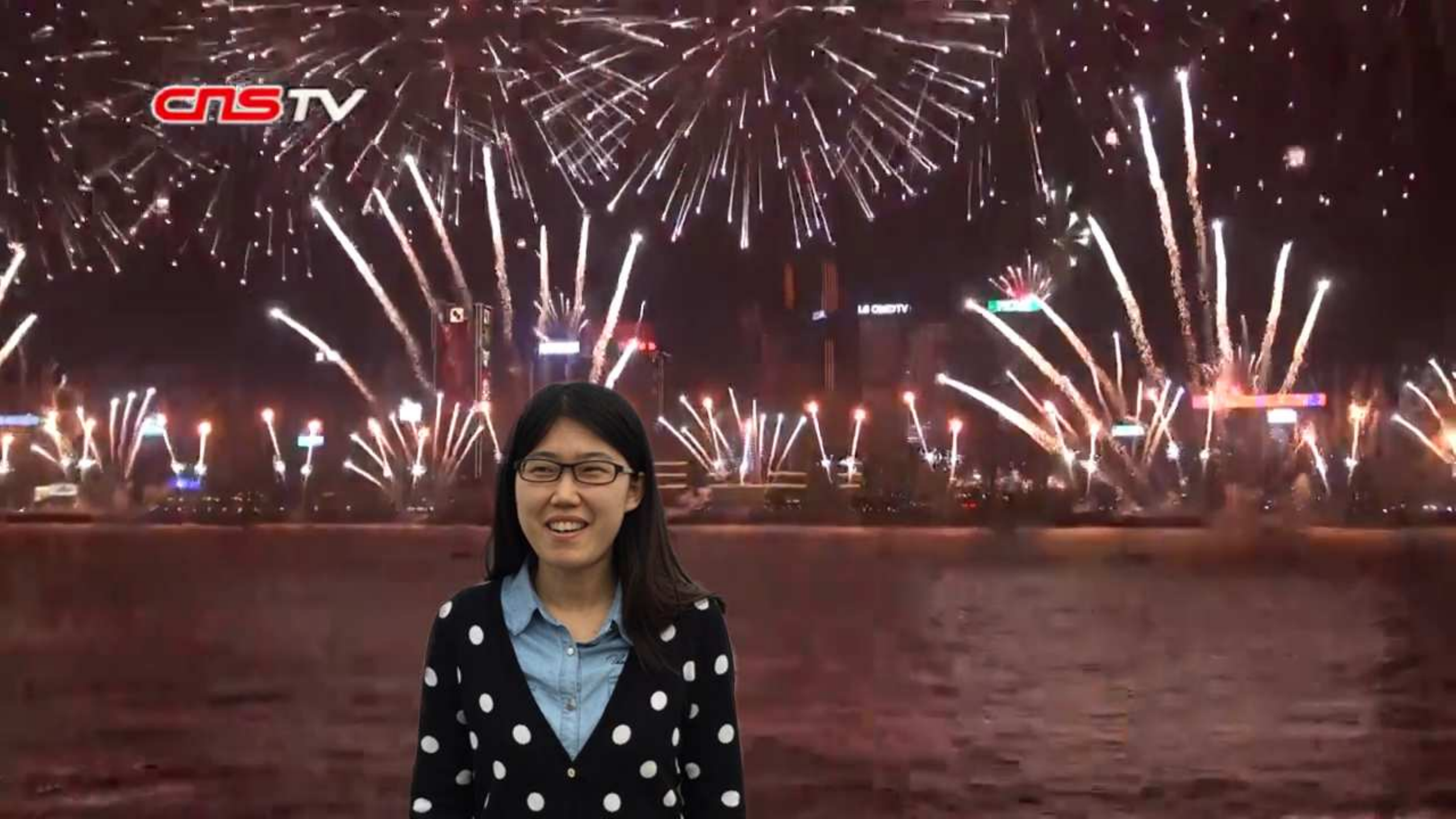}&
\includegraphics[width=0.19\linewidth]{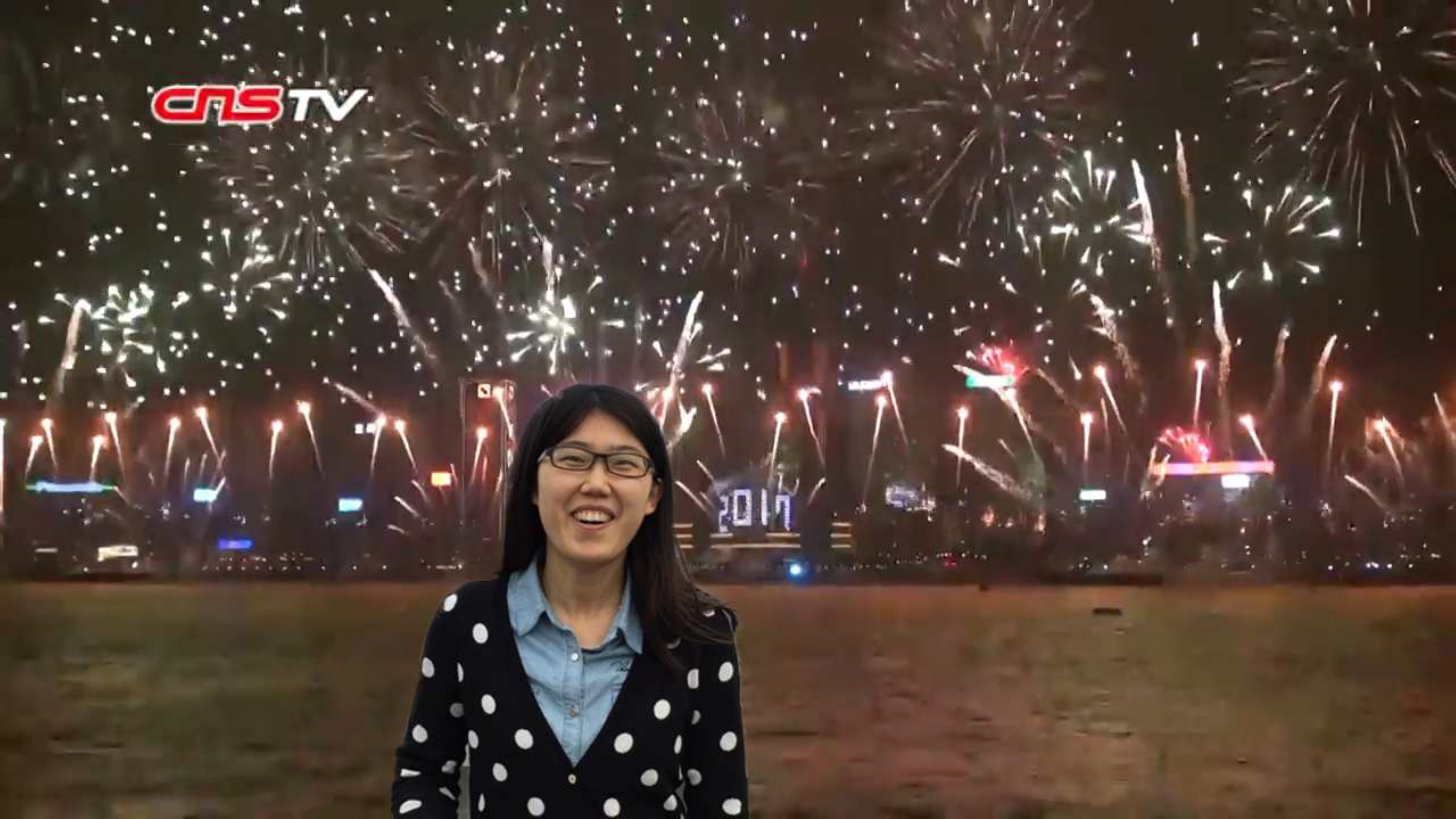}&
\includegraphics[width=0.19\linewidth]{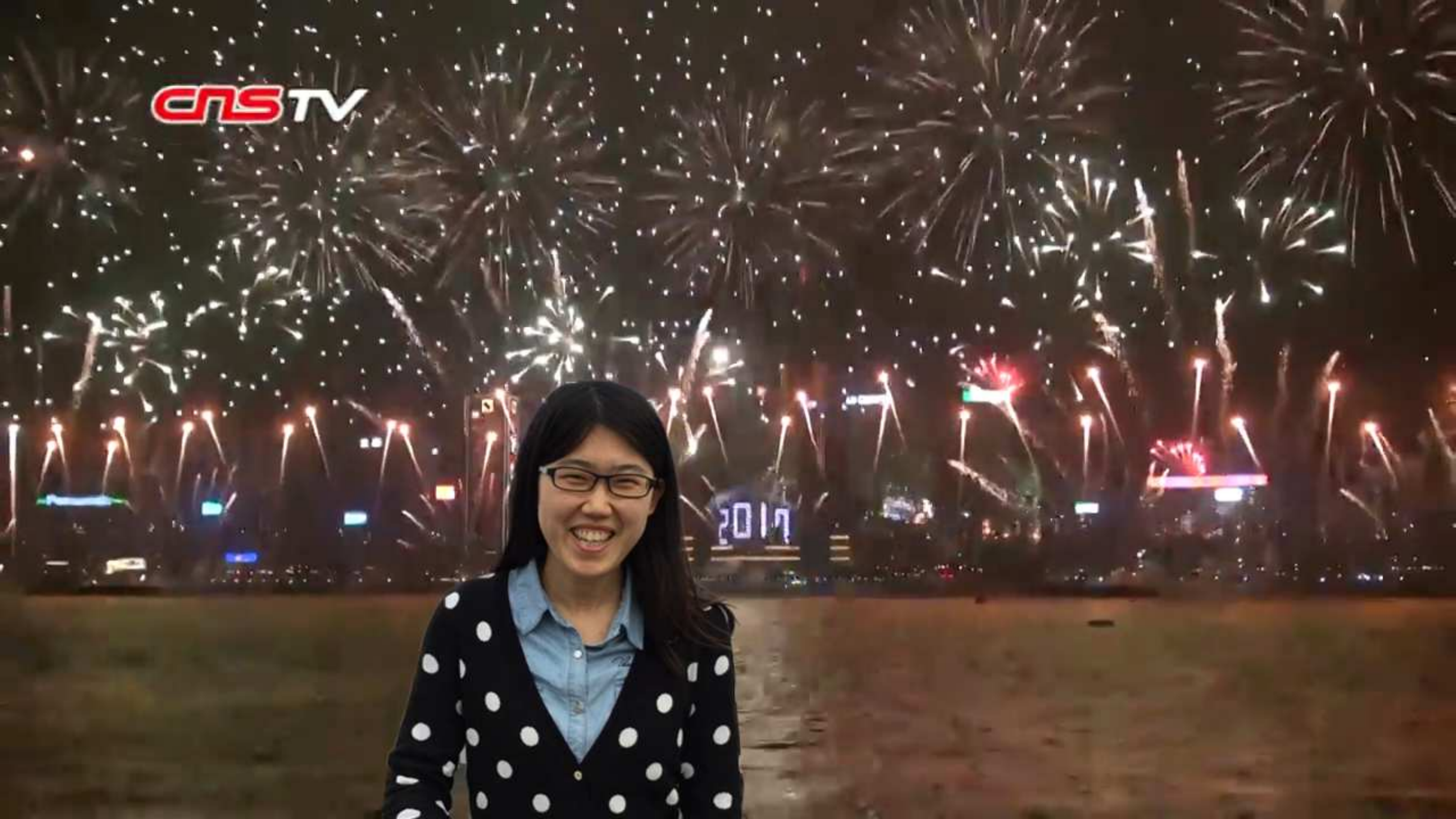}&
\includegraphics[width=0.19\linewidth]{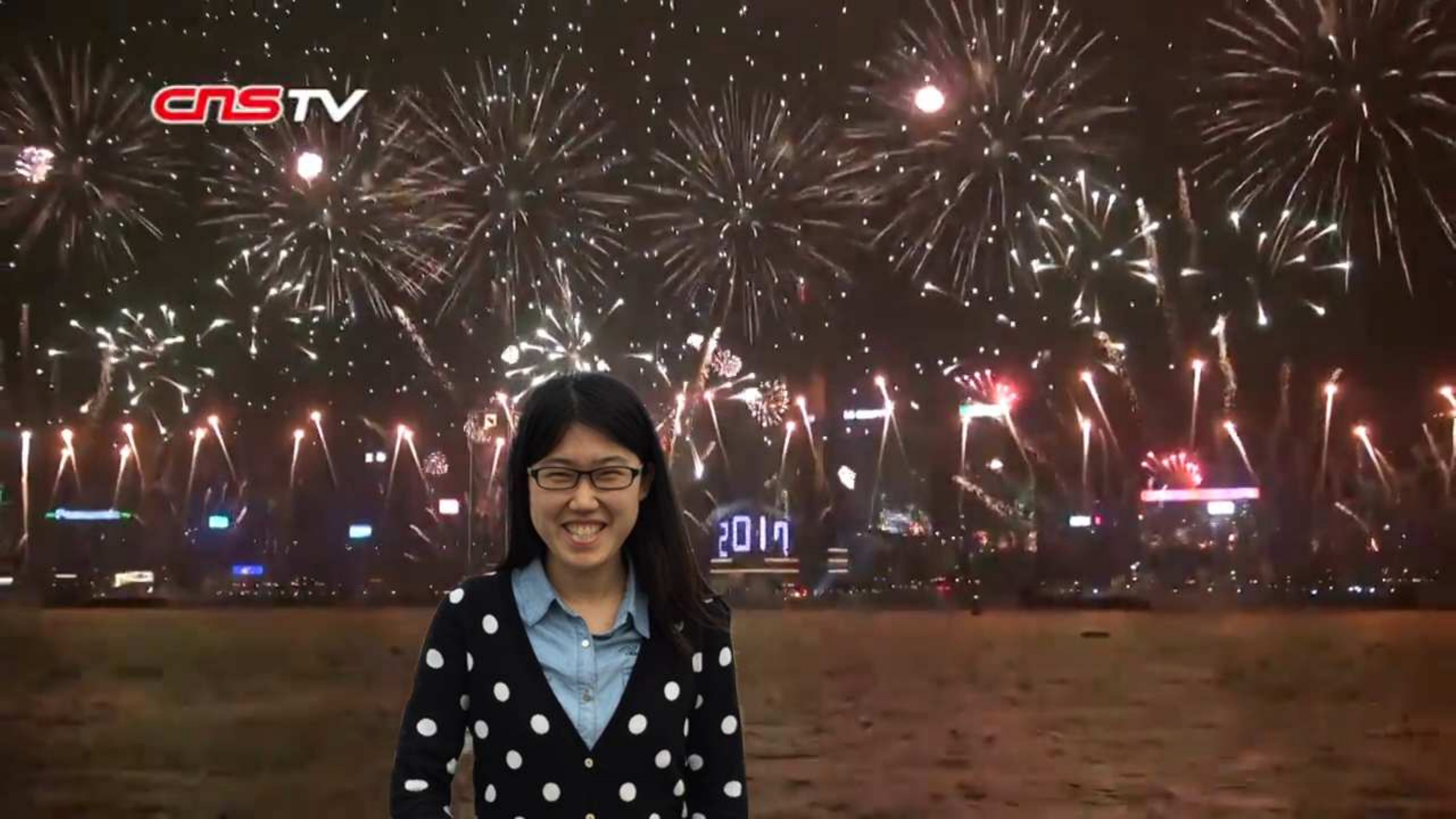}\\
\end{tabular}
\caption{Applications of real-time background editing. The top row shows the captured
videos by iPhone 7 and the corresponding segmentation maps by our method. The bottom row
is the blending results on new background. } \label{fig:application}
\end{figure*}

\vspace{0.1in}\noindent\textbf{Visual Comparisons~~} We show more comparisons in Figures
\ref{fig:comparison1} and \ref{fig:comparison2}. The two examples are complex scenes. FCN
\cite{long2015fully}, DeepLab \cite{Chen2014_deeplab} and PSPNet \cite{ZhaoSQWJ16}
results shown in (b-d) cannot well distinguish between foreground and background in pixel
level. Ours with background attenuation can well cut out foreground. Background samples
we apply for attenuation is the background of the first frame. As shown in (e), the
spatial-temporal refinement further improves boundary accuracy. More results are shown in
our supplementary material.

\subsection{Other Applications}
Our method provides the suitable solution to edit portrait video background in real-time.
In Figure \ref{fig:application}, we show an example for automatically video background
change. With the high-quality segmentation map, the portrait can be blended into new
background images. More applications such as video stylization, depth-of-field are
exhibited in our supplementary material.

\section{Conclusion}
We have presented the automatic real-time background cut method for portrait videos. The
segmentation quality is greatly improved by the deep background attenuation model and
spatial-temporal refinement. The limitations of our methods are that our approach may
fail for severe blur caused by fast motion and very dynamic background. Addressing these
limitations will be our future work.

{\small
\bibliographystyle{ieee}
\bibliography{deep_bg_seg}
}

\end{document}